\documentclass[journal]{IEEEtran}
\usepackage[english]{babel}             
\usepackage[
    backend=bibtex, 
    style=numeric,
    sorting=none, 
    maxnames=50
]{biblatex}                             
\usepackage{csquotes}                   
\usepackage{graphicx}                   
\usepackage{hyperref}                   
\usepackage[
    user,
    titleref
]{zref}

\title{Reservoir Computing in robotics: a review}
\author{
    Paolo Baldini
    \\
    \href{mailto:paolo.baldini6@studio.unibo.it}{paolo.baldini6@studio.unibo.it}
}
\date{\today}

\begin{document}

\maketitle

\begin{abstract}
    Reservoir Computing is a relatively new framework created to allow the usage of powerful but complex systems as computational mediums.
    The basic approach consists in training only a readout layer, exploiting the innate separation and transformation provided by the previous, untrained system.
    This approach has shown to possess great computational capabilities and is successfully used to achieve many tasks.
    This review aims to represent the current `state-of-the-art' of the usage of Reservoir Computing techniques in the robotic field.
    An introductory description of the framework and its implementations is initially given.
    Subsequently, a summary of interesting applications, approaches, and solutions is presented and discussed.
    Considerations, ideas and possible future developments are proposed in the explanation.
\end{abstract}

\section{Introduction}
\zlabel{section:introduction}
Learning techniques are widely used in robotics with the aim of automatising the generation of control software, from parameter tuning to the whole behaviour.
The robot should automatically learn new tasks or behaviours, possibly allowing the programmer to abstract away from low-level problems and tunings.
Nevertheless, the choice of some actions requires analysis that are related to space and time.
Tasks like generation of gait patterns, for example, require the system to know the previously generated values, basically requiring some sort of memory to be present.
This problem may be addressed with the creation of feedback links in a neural network, allowing it to generate the required function.
Depending on the task complexity however, this is not always feasible.
Some behaviours require different temporal information that might not be obvious during the system creation and set up.

Recurrent Neural Networks represent a promising way to approach those problems, embedding recurrent links and allowing the analysis of temporal data.
These systems, even if potentially very powerful, bring the drawback of a complex learning, that is indeed time and power consuming and makes the training really expensive.
Training and computational capabilities of Artificial Networks are not the only problems in robotics usage.
The used network size may represent a problem for small robots.
Indeed, their available memory may not be enough for the storage of complex models.
Furthermore, the complexity of the input/output processing may slow down the system in case of massive networks and reduced computing power.
This kind of problems are obviously negligible for many types of robots, but this is not the case, for example, of robots controlled through use of microcontrollers.

All the described problems make the development of powerful robotic systems harder and slower.
However, a relatively new framework known as Reservoir Computing may possess the characteristics needed to solve or mitigate some of these complications.

\subsection{Reservoir Computing}
The aim of the Reservoir Computing framework is to allow the exploitation of powerful systems that are however hard to train or configure.
In some cases, they may be also impossible to modify without changing their intrinsic aspects.
This characteristic would normally prevent their usage as computing systems, in that they cannot be easily tuned or configured for the specific task.
Reservoir Computing allow their usage, bypassing the training problem.
In the framework, those systems are called `reservoirs'.
The final system that exploit their `raw power' is called `Reservoir Computing'.

Reservoir systems can be of different types: Recurrent Neural Networks; soft bodies; biological systems; physical structures and so on.
Those have high computational power, but each of them may express it differently.
A clear example is provided by the reservoirs used in \cite{PatternRecognitionInABucket} and \cite{MovementGenerationAndControlWithGenericNeuralMicrocircuits}.
In the first case, a bucket of water is used; in the latter one, the reservoir is just an artificial network.
Both systems embed computational abilities, but they do it respectively using a water medium and a recurrent neural network.
Those differences make it hard to define what a reservoir is, unless through the definition of the characteristics that it must have.

The first requirement is the ability to project the input data to a higher dimensional space, performing a separation of the input signal.
The resulting state has to be different for different input series, such as it can be used to discriminate different sequences of incoming data.
This property is called `separation property'.
Finally, the system state should depend on the history of the inputs, but the influence of the older values should decrease in time.
This property, that is not present in chaotic networks \cite{RealTimeComputationAtTheEdgeOfChaosInRecurrentNeuralNetworks}, is called `echo state property' \cite{TheEchoStateApproachToAnalysingAndTrainingRecurrentNeuralNetworks} or `fading memory' \cite{RealTimeComputingWithoutStableStatesANewFrameworkForNeuralComputationBasedOnPerturbations}.
Those requirements are what make the reservoir different from classical feed-forward networks and allow the framework to achieve, in some cases, higher results in its usage.
Nevertheless, the concept of Reservoir Computing has evolved with time, and some different points of view gained popularity, slightly changing some basic ideas.
In this review, the differences in those approaches are addressed as `variants' of the original idea, and are later explained (section `\ztitleref{section:architectures}').

The non-linear transformation of inputs with the fading memory property allows to obtain spatio-temporal output patterns.
We can say that those are directly extracted from the incoming sequence of data, in that they change depending on the previous values.
The patterns originated from the input sequence can be utilised for the training of a readout (see figure \ref{figure:reservoir_architectures}A).
This is usually a network trained with a simple algorithm like linear regression or classification.

To conclude, the advantage of Reservoir Computing consists in the decrease of complexity in the task of training the network.
The learning in the readout is indeed much easier compared to the one of a recurrent system.
Indeed, this approach also completely avoids the problems of training a network for dealing with temporal data, in that it is instead performed by the reservoir system.
This allows to exploit the advantages provided by a Recurrent Neural Network with just the need of training the readout with an easier method.

To sum up, Reservoir Computing implicitly resolves some of the previously discussed problems.
It allows faster training in that it just requires the training of the readout.
This consequently brings a reduction in power consumption, that may be further reduced if specific kinds of reservoirs are used.
Some examples are electrical-circuits or photonic reservoirs.
Also the load on the CPU is reduced, needing an easier training and possibly allowing a `distributed' computation, that moves from the CPU to the reservoir.
Finally, it allows to perform time-dependent analysis of the input data, removing the need to train a Recurrent Neural Network or set up ad-hoc designs of the system.

\subsection{Reservoir computing architectures}\zlabel{section:architectures}
Reservoir Computing is a fairly general framework, and thus many variants exist.
Here the most interesting ones are briefly overviewed, explaining their main concepts and why they can be considered as variations of the original framework.
The type of the reservoir used will not be discussed or taken in consideration if it is not a fundamental part of the architecture.

\begin{figure*}\centering
    \includegraphics[width=.9\textwidth]{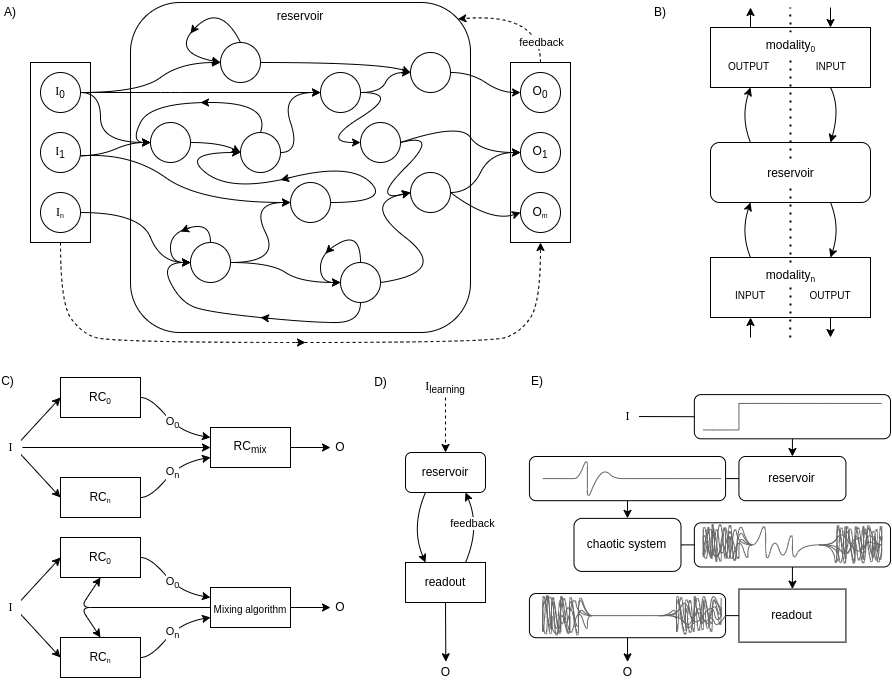}
    \caption{
        Architectures of the Reservoir Computing frameworks and its variants.
        The dotted lines represent optional or temporary connections.
        A) The Echo State Network idea.
        All the other Reservoir Computing proposals follow the basic principles of taking input data, elaborate them through the non-linear reservoir and evaluate its state with a simple method.
        B) A schematization of the multi-directional system presented in \cite{MultiDirectionalContinuousAssociationWithOnputDrivenNeuralDynamics}.
        The modalities can be used both as input and output.
        Settings of values in a modality will force the others to adapt to the changes.
        C) The MACOP (bottom) and the modular/hierarchical architecture presented in \cite{ModularReservoirComputingNetworksForImitationLearningOfMultipleRobotBehaviors} (top).
        As visible, the main difference consists in the mixing method.
        D) An input-less system.
        The learning input is usually used during the training, while it is substituted by feedback loop when the system start to behave correctly.
        E) The architecture exploiting a chaotic neural network, presented in \cite{SoftBodiesAsInputReservoirRoleOfSoftnessFromTheViewpointOfReservoirComputing}.
        As visible, the input signal causes a perturbation in the reservoir.
        This induces a temporary stabilization of the chaotic signal.
        The readout basically separates the stable behaviour from the chaotic one, allowing for a temporal analysis.
    }
    \label{figure:reservoir_architectures}
\end{figure*}

Starting with one of the most different variants, in \cite{MultiDirectionalContinuousAssociationWithOnputDrivenNeuralDynamics} the authors propose a multidirectional architecture (see figure \ref{figure:reservoir_architectures}B).
Here, the concepts of input network and readout are mixed in what is called `modality'.
This acts like an interface to the underlying reservoir system.
The modalities change their role depending on the way they are used, basically acting as output if another modality is used as input.
This conception is interesting in that it allows to generate a bidirectional Reservoir Computing system, that can take multiple types of input and generate the correct output.
Nevertheless, this solution is harder to implement compared to a classical reservoir system and may be interesting just in some specific applications.

A hierarchical approach in the usage of reservoir networks is presented in \cite{MACOPModularArchitectureWithControlPrimitives} and \cite{ModularReservoirComputingNetworksForImitationLearningOfMultipleRobotBehaviors}.
Both propose the usage of multiple modular Reservoir Computing frameworks, accurately `mixed' for the achievement of a final goal (see figure \ref{figure:reservoir_architectures}C).
The differences consist in the concept behind those modules and in the way of blending their results.
The first implements the sub-controllers with the aim to solve `control primitives'.
The mixing is done by an external system able to assign the responsibility of achieving a specific action.
The second instead defines sub-controllers able to achieve sub-tasks like object following and obstacle avoidance.
The mixing is finally done by a larger reservoir, able to choose the correct action.
The power of these architectures is claimed by both the group of authors, and may follow the well-know principle of \textit{divide et impera}.
Nevertheless, the complexity in the training of those systems seems higher compared to more classical Reservoir Computing solutions.

Unconventional type of Reservoir Computing variant may be also recognised in input-less systems (see figure \ref{figure:reservoir_architectures}D).
Those are often used for pattern generations and do not include any controllable input.
Indeed, they often only have a trained feedback from the output of the readout to make the oscillation proceed.
However, the lack of explicit inputs does not completely exclude their presence.
Morphological computing bodies used as reservoir are often used for pattern generation without inputs.
They are however intrinsically embedded in the environment and their computation and state depend on it \cite{EnvironmentalandStructuralEffectsOnPhysicalReservoirComputingWithTensegrity}.
This influence may be seen as an input from the space around the robot, also allowing to exploit the system for sensing \cite{TheBodyAsAReservoirLocomotionAndSensingWithLinearFeedback}.
The acceptance of this architecture as a class of Reservoir Computing is controversial, especially given that one of the base concept of the framework is to perform a transformation of the input to a higher-dimensional space.
Nevertheless, the feedback and the environmental interaction may be seen as inputs to the system, allowing this interpretation to be accepted.
Moreover, the concept of exploiting an untrained system is still obviously supported.

Finally, one of the most particular variants of the Reservoir Computing framework is presented in \cite{SoftBodiesAsInputReservoirRoleOfSoftnessFromTheViewpointOfReservoirComputing}.
The concept of this architecture consists in the exploitation of a chaotic neural network for time dependent tasks (see figure \ref{figure:reservoir_architectures}E).
The system is composed of a normal (i.e., ordered or critical) reservoir used as `input layer'.
Its raw output is then injected into a chaotic system, whose state is read by the readout.
The idea is to exploit an `innate trajectory' defined by the chaotic dynamics of the network and strengthened by a specific algorithm.
An input will then induce the trajectory to stabilise for some time.
An analysis of its state is then performed by the readout, that can exploit it for time-dependent tasks, like usage as a timer.
The validity of this variant seems limited to a specific niche of problems; nevertheless, a confirmation of its capabilities may bring to an increase in the performance of the related solutions.

\subsection{Types of reservoir}
Although many `exotic' types of reservoir exist \cite{ReservoirComputingApproachesToRecurrentNeuralNetworkTraining}, just few of them seem to be effectively used in practical applications.
Therefore, we will identify and describe some macro-types of reservoirs actually used in robotic works, or the ones that seem more interesting for them.
Later, we will examine them in-depth, providing examples of their usage and results.

As said, Reservoir Computing just needs a system that provides a non-linear, high dimensionality input projection with fading memory.
Those characteristics concur in the extraction of important spatio-temporal features from the incoming sequence of data.
Multiple systems provide those characteristics, allowing them to be used as reservoirs.
The original approaches that unified into Reservoir Computing are Echo State Networks \cite{TheEchoStateApproachToAnalysingAndTrainingRecurrentNeuralNetworks} and Liquid States Machines \cite{RealTimeComputingWithoutStableStatesANewFrameworkForNeuralComputationBasedOnPerturbations}.
In those systems, the non-linear component is an untrained Artificial Neural Network: a Recurrent Neural Network for the first; a Spiking Neural Network for the second.
However, any type of system that provides the required characteristics may in theory be used.
This is shown in some experiments like \cite{PatternRecognitionInABucket}, where a bucket of water is used as a reservoir.

Given the flexibility of the framework, a distinction of the multiple types of reservoirs is useful for a better comprehension of the topic.
The categorization actually present in the literature is however very specific, possibility making this overview more complex than it should.
Therefore, an arbitrary classification in some macro-categories is performed, grouping different types of reservoirs according to some common characteristics.
The categories will refer to the reservoirs used in robotic studies until now.
Because of that, some other solutions may not be analysed.
Moreover, specific categories will be described only when they add useful information to the study.

Based on the analysis of the literature, a good distinction of reservoirs is given by the following categories:
\begin{itemize}
    \item biological systems
    \item physical networks and systems
    \item morphological computing bodies
    \item software networks
\end{itemize}

We can see that one of the categories is `morphological computing bodies'.
This may be considered a subset of `physical systems'.
However, the fundamental ideas of the two classes are too different to be ignored.
If a physical system may be an addition to the robot, the body is instead an integral part of it.
Following this reasoning, a distinction of the two categories is decided.

Regarding `biological systems', those might be considered as `physical systems' as well.
However, the label `physical' is here used only for reservoirs composed by hardware components, thus excluding living systems.

A reader might object that a robot body might be in theory composed of biological components, suggesting the grouping of `biological systems' and `morphological computing bodies' to a single category.
This objection is however distant from reality in that, to the best knowledge of the author, no useful result uses this type of system.
Although this objection may become valid in future, the actual distinction is strong enough for the goal of this review.

\subsection*{Outline}
This paper begins providing an introductory description of the Reservoir Computing framework and its derived architectures.
Subsequently, a types-set for the classification of reservoir systems is proposed.
This identifies the following classes:
\begin{itemize}
    \item biological system;
    \item physical networks and systems;
    \item morphological computing bodies;
    \item software networks.
\end{itemize}
The aim is to highlight common properties and characteristics of the systems, together with the approaches that are more suitable for their usage.
The proposed categories are detailed and motivated in specific sections (see sections `\ztitleref{section:biological}', `\ztitleref{section:physical}', `\ztitleref{section:morphological}', `\ztitleref{section:software}').
Also, examples of systems belonging to these classes are provided.

Finally, the most interesting results and solutions from articles are presented and discussed (see section `\ztitleref{section:applications}').
During the analysis, points to ponder and ideas are proposed.

\section{Biological systems}
\zlabel{section:biological}
Biological systems show computational abilities that are hard if not impossible for the actual technologies to reproduce.
The human brain has billions of neurons and a much greater number of synapses.
Exploiting such complex networks may lead to an incredible increment in computing capabilities of systems.
Biological networks naturally embed recurrent links, making them adapt for the analysis of spatio-temporal data-series.
Moreover, they possess neural plasticity, a property allowing complex adaptation of the network through modification of connections and growth of the system.

Although powerful, a downside of those systems is their being hard to exploit.
Their growth cannot be completely controlled, making their adaptation to a specific task complex.
Moreover, the interconnection between biological systems and electrical components is not straightforward, and it is hard to set up.
Despite those limitations, researchers are actively trying to exploit biological neuronal systems as computation medium.
A promising way seems to be related to the usage of Reservoir Computing.
The structure of the framework effectively allows to use the biological network as the reservoir, exploiting its ability in spatio-temporal dependent analysis.
This methodology is also used in robotics, where those systems are used for tasks achievement \cite{ReservoirComputingWithDissociatedNeuronalCulture} \cite{TowardsMakingACyborgACloseLoopReservoirNeuroSystem}.

The actual state-of-the-art in usage of biological systems as reservoirs exploits cultures of neuronal cells from different origins, like from rats.
It is proven that those possess the echo state property and can thus be used in the Reservoir Computing framework \cite{EchoStatePropertyOfNeuronalCellCultures}.
Alternatives proposed the usage of bacteria populations \cite{IsThereALiquidStateMachineInTheBacteriumEscherichiaColi}.
This latter one however has not been used for robot control.

Whatever the system is, a medium for its interaction with the software is still needed.
Technological solutions address this problem and set up an electrical communication between the two worlds.
This allows biological systems to be used as reservoirs.
However, these technologies does not resolve the problem of the coding and decoding of the information, that still remains a difficult point in the exploitation of those systems.

A common way of interfacing the biological networks to be used as reservoirs consists in the usage of a Micro-Electrodes Arrays, shortly called MEA.
Those stimulate the extracellular potential of biological neuronal cells, generating a downstream activation of the network nodes.
Other possibilities for the neurons' stimulation include use of chemical components like dopamine \cite{TowardsMakingACyborgACloseLoopReservoirNeuroSystem}.
The activation will eventually propagate to the groups of neurons selected as output, whose values will be then recorded and used as input for the readout.
Because of physical limitations, the sensors at the output of the biological network perceive a group of neighbour neurons instead of a single one.
The simplest possibility for output spikes detection thus consists in considering the entire group as the output node, instead of the single neuronal cell.
This obviously reduces the information obtainable from the biological network, in that the single activation is no longer visible.
Nevertheless, spikes in a neighbourhood seem to be related \cite{TheHippocampusAsASpatialMapPreliminaryEvidenceFromUnitActivityInTheFreelyMovingRat}, possibly not compromising too much the final result.
Moreover, methodologies for single spike detection from noisy signals exist in literature \cite{AReviewOfMethodsForSpikeSortingTheDetectionAndClassificationOfNeuralActionPotentials} and may allow for a finer output analysis.

The recorded output is used as the extracted pattern and fed to the readout.
The idea of using a machine learning technique is presented in \cite{ControllingAMobileRobotWithABiologicalBrain}, where the authors proposed its usage for the evaluation of the output-to-actuators transformation of the responses of the biological system.
This approach is fully represented by Reservoir Computing techniques and may be used to train the output networks to achieve tasks.
Examples may be the direct control of robot actuators (e.g., motors) as well as process information from sensors.

\section{Physical networks and systems}
\zlabel{section:physical}
A reservoir is a component of the Reservoir Computing framework with the ability of a non-linear, history dependent input analysis.
Its output represents a spatio-temporal pattern that can be used by a trained readout for achieving a specific goal.
The given description does not restrict the reservoir to a network system, but allows a variety of different mediums to be used.
To address reservoirs that are not software, biological, or directly belonging to the robot body, the label `physical systems' is chosen.
A physical system may be a hardware implementation of a neural network, as well as any kind of material complex system that is able to provide the required computational capabilities.

The literature presents many studies about physical reservoirs.
Implementations comprehend analogical electrical circuits \cite{InformationProcessingUsingASingleDynamicalNodeAsComplexSystem}; Field-Programmable Gate Arrays (FPGAs) \cite{AHardwareOrientedEchoStateNetworkAndItsFPGAImplementation}; Application-Specific Integrated Circuit (ASIC) in a Very Large-Scale Integration (VLSI) fashion \cite{EdgeOfChaosComputationInMixedModeVLSIAHardLiquid}; memristive components \cite{MemristorBasedTeservoirComputing}; photonic \cite{TowardOpticalSignalProcessingUsingPhotonicReservoirComputing} and spintronic \cite{NeuromorphicComputingWithNanoscaleSpintronicOscillators} reservoirs; mass-spring systems.
This latter ones are however often related with morphological computation.

Despite the many implementations available, their application in robotics is still at its first steps.
Few studies like \cite{AHardwareOrientedEchoStateNetworkAndItsFPGAImplementation} propose usage of those physical implementations in robotics, however, to the best of the author's knowledge, no study has been effectively conducted on their usage.

Physical reservoirs represent an interesting development for the robotic research.
Although not as configurable as software systems, they usually present much higher performances in comparison.
Particularly, the battery consumption proved to be way lower and the processing speed much higher \cite{AHardwareOrientedEchoStateNetworkAndItsFPGAImplementation}.
Moreover, ad-hoc components produced in large quantities might be cheaper than general purpose computational systems.

\section{Morphological computing bodies}
\zlabel{section:morphological}
To talk about morphology in its usage as a computational medium for the Reservoir Computing framework, a propaedeutic definition of `morphological computation' is needed.
Here, it is defined as a kind of advantageous information processing performed by the body of the robot itself \cite{WhatIsMorphologicalComputationOnHowTheBodyContributesToCognitionAndControl}.
This definition assumes that the system performs some operations, and thus excludes the ones that simply `enable' or `simplify' the computation.
Moreover, the body must be able to properly transform general types of encoded inputs.
This excludes all the systems that do not allow generic data to be analysed.
Nevertheless, it is important to highlight that the presented studies do not always exploit this ability to compute general information.
This does not invalidate the chosen definition, but allows to accept system implementations and architectures that simply do not exploit this advantage (e.g., CPG).

Usually a body cannot be considered a computing machine, in that it cannot perform calculations with given, general, encoded inputs.
Nevertheless, new kinds of robotic bodies are actually opening the way for this kind of data processing.
Soft bodies have the characteristic of adapting to the environment instead of resisting to it.
They also possess high dynamics in that their behaviour depends on the input, the environment, and the actual state of the body itself.
This dependence naturally embeds also a concept of fading-memory, in that the actual state of the system depends on its evolution in time.
It is then possible to theoretically encode any type of input and feed it to the soft body in the form of stimulation.
The system will respond to the input and its final state will be then recognised as a calculated output.
This behaviour theoretically allows to offload part of the computation to the robot itself.

Multiple studies evaluate the computational capabilities of soft bodies \cite{ASoftBodyAsAReservoirCaseStudiesInADynamicModelOfOctopusInspiredSoftRoboticArm} \cite{ExploitingShortTermMemoryInSoftBodyDynamicsAsAComputationalResource}.
Those show the potential of this type of systems, however also highlighting some weak points.
The first consists in their low processing speed.
The input signal is usually given to the system through a perturbation of the state, for example using a motor.
This stimulation should have time to propagate in the body before being useful for a state read.
This spreading of data is often very slow and may limit the computational abilities of the system.
Moreover, the state is often read through usage of sensors embedded in the body which, due to their cost and space, cannot usually be in a high quantity.
This problem is tackled in \cite{ComputingWithAMuscularHydrostatSystem} and \cite{ExploitingTheDynamicsOfSoftMaterialsForMachineLearning}, where a technique for virtual augmentation of output neurons is proposed.
This consists in using a different timescale for input injection and output readings, allowing to collect the output value multiple times before the new input is supplied.
The added value of this technique is that it is possible to better perceive the response of the body to the stimulation in the time context.
This allows to better understand the response of the system by the readout.
The results showed an effective increase of the computational power, often allowing the system to score as well as other systems with as many real neurons as the virtualised ones.
Although this technique is not granted to improve the performance in every task, its usage may be useful for achieving some goals that the normal system is not already able to achieve.
Few output nodes are however enough for many useful tasks like pattern generation \cite{MorphologicalPropertiesOfMassSpringNetworksForOptimalLocomotionLearning}.

Alternatives to sensors for the perception of the state of the system exist, and consist in the use of cameras or the body itself.
These might theoretically resolve the problem of a limited number of information.
Nevertheless, their usage is not as straightforward or reliable as usual sensors.
A proposal for state detection through image acquisition is presented in \cite{FromTheOctopusToSoftRobotControlAnOctopusInspiredBehaviourControlArchitectureForSoftRobots}.
The approach is used for the emulation of the natural octopus behaviour, changing the action when an object is detected \cite{OnlineLearningForBehaviorSwitchingInASoftRoboticArm}.
It is also used to detect the deformation of the soft body in a sensor-less arm \cite{SensingThroughTheBodyNonContactObjectLocalisationUsingMorphologicalComputation}.
The goal is to use the system to detect the position of an object artificially-invisible to the camera.
Another proposal for the state detection is presented in \cite{EmulatingASensorUsingSoftMaterialDynamicsAReservoirComputingApproachToPneumaticArtificialMuscle}, where current traversing a Pneumatic Artificial Muscle and depending on its state (relax/stiff and load) is successfully used as input for a reservoir.
This use depends on the modification of the body with the embedding of carbon particles.
The aim is to make the system conductible and thus able to sense its own state.

Nevertheless, it has to be noted that, also if soft bodies are able to compute general information, it may not be worth to use them for generic operations.
Since they have to deal with the environment, using them as computational medium means requiring multiple computations to be possibly carried on contemporarily.
Moreover, differently from other type of reservoirs, soft bodies are usually slow in processing.
Due to these characteristics, the computational power provided by those systems may not be enough for some tasks.
Given that a proper evaluation method does not actually exist, the only possibility is using heuristics or directly testing the system.
This approach risks however to greatly slow down the development of the robot.
On the other hand, their being part of the body means that no additional system has to be added, making them more appealing to distribute the computations and for updates of the control system.

Their usage is useful for the implementation of Central Pattern Generators directly embedded into the body, closely mimicking biological beings.
The idea of offloading the computation in the body comes indeed from the natural world.
Biological entities like humans \cite{CentralPatternGenerator2000} \cite{CentralPatternGenerator2015} and octopus \cite{FromTheOctopusToSoftRobotControlAnOctopusInspiredBehaviourControlArchitectureForSoftRobots} embed part of their control to the body itself.
This offloading is often aimed at the generation of movement patterns that control appendages and limbs.

Having discussed the characteristics of soft bodies, it is possible to understand their utility and potential in usage as reservoirs.
The idea of the framework consists in a sequence of related data to be analysed by a system able to identify spatio-temporal patterns.
This computation is provided by the soft body thanks to its high dynamics and fading-memory property.
The system is then able to extract the pattern that will be used as input for a trained readout.
In other words, applying Reservoir Computing to soft body mediums consists in learning the response of the body to given sequences of data.
This can be done by simply embedding sensors in it.

The development of soft bodies brings multiple benefits.
It is essential in that it allows usage of robots in human environment, adapting to it and causing less harm compared to normal hard bodies.
Beside of that it also allows computation to be carried on without the need of additional computational systems, possibly making the robot cheaper.

\section{Software networks}
\zlabel{section:software}
The `software networks' discussed in this paper are a specific type of Artificial Neural Network defined in software.
This distinction allows us to differentiate this type of computational medium from other systems that have their nodes defined in a more static way.

This distinction is important in that it completely changes the approach to those kinds of systems.
For static or hardly modifiable networks, Reservoir Computing is one of the few ways to exploit their potential.
On the other hand, software neural networks are usually highly configurable and may be used with good results also without the need of the framework.

Their configuration flexibility comes however with a cost.
Software neural networks usually exploit usage of CPUs or GPUs, that makes them slower than ad-hoc hardware implemented networks.
Electrical circuits or photonic systems, for example, exhibit far better performance compared to them, at least when the signals propagation is considered.
Those hardware implementations usually consume less electrical power compared to software neural networks: a characteristic very important in robotics.
Finally, some other computational mediums, like biological ones, might have more potential processing power compared to them.

Despite their non-optimality, the flexibility of software neural networks make them the most used type between the previously described systems; in both computer science and robotics.
Moreover, they are fast and easy to create, making them optimal for tests and studies.
Finally, not all the applications require low-power consumption or high computational abilities, and also if they do, modern CPUs and GPUs are usually powerful enough for most of the tasks.

Because of their characteristics, specific types of software neural networks that provide high spatio-temporal input separation are heavily studied as reservoirs.
Specifically, Echo State Networks and Liquid State Machines, the two framework from which Reservoir Computing originates, were initially implemented with software networks as reservoir, respectively a Recurrent Neural Network and a Spiking Neural Network.

As for other systems used as reservoir, software networks need spatio-temporal analysis capability.
This property is usually given by their recurrent links, allowing for an echo state property and thus a capability in analysis of correlated data sequences.
Their fading memory characteristic may then produce spatio-temporal output patterns, depending on previously seen data.
In other words, the computation must depend on the state/context of the network, that directly depends on its history.
The described characteristics allow those networks to be used as reservoirs.
This trend is also visible in the robotic field, where many experiments use those type of networks.

In the following sub-sections, we will briefly describe the most common types of software neural networks used as reservoirs.


\subsection{Spiking Neural Networks}
Spiking Neural Networks try to mimic natural network models.
To clearly describe them, we have to give an overview on the functioning of biological systems.

First, natural neural networks do not work in a continuous way, but with inputs and outputs encoded in pulses.
This model requires the neurons to be different from the ones used in other types of artificial systems.
The pulses spread through the synapses and charge the potential of the membrane of the neurons that they reach.
The potential increases until it reaches a specific level, at which it is able to pass through the membrane and proceed to the output synapses.
The signal is then spread to other neurons, charging them and possibly generating downstream propagation through multiple neurons.

Biological systems heavily influenced the development of artificial networks, with Spiking Neural Networks that closely try to mimic the natural model of neurons and, in some implementations, synapses \cite{RealTimeComputingWithoutStableStatesANewFrameworkForNeuralComputationBasedOnPerturbations}.
Moreover, natural networks include recurrent connections, allowing inspired models to be used for time related analysis.

As we said, in Spiking Neural Networks the inputs and the outputs are in the form of spike sequences.
This characteristic make them more complicated to use compared to first or second generation of neural networks.
Each input has indeed to be represented through pulses instead, for example, of analogue values.
Many solutions have been provided for this conversion, often requiring some sort of time elaboration \cite{NetworksOfSpikingNeuronsTheThirdGenerationOfNeuralNetworkModels}.
Alternatives involved the injection of the analogical inputs through a spatial representation \cite{MovementGenerationAndControlWithGenericNeuralMicrocircuits}.
As to compensate for their difficult approach, it is proven that Spiking Neural Networks are more powerful than same size networks of previous generations \cite{NetworksOfSpikingNeuronsTheThirdGenerationOfNeuralNetworkModels}.

\subsection{Recurrent Neural Networks}
Recurrent Neural Networks are artificial networks that take inspiration from the recurrent links present in biological systems.
This characteristic allows them to deal with sequences of data, exhibiting a temporal analysis ability.

Derived from Feedforward Neural Networks, Recurrent Neural Networks can natively work with continuous data, allowing to exploit temporal analysis without having to deal with the spikes generated by a Spiking Neural Networks.

This type of networks are the ones originally used in Echo State Networks \cite{TheEchoStateApproachToAnalysingAndTrainingRecurrentNeuralNetworks} and the most used ones in Reservoir Computing studies.

\subsection{Boolean Neural Networks}
A special type of Artificial Neural Networks is the one using boolean algebra as its base.
Its main characteristic is the usage of binary values in the system's state and boolean operations as the nodes' functions.
Those networks are interesting in that they can be used for efficient and computationally easy calculations.
Because of that, their usage in simple robots is more practical compared to other types of software neural networks.
Nevertheless, their usage in Reservoir Computing, especially in robotics, is not common and not well evaluated.
A recent publication analysed their usage in the framework, showing that they can be successfully exploited for the emulation of recursive binary filters \cite{FlexibilityOfBooleanNetworkReservoirComputersInApproximatingArbitraryRecursiveAndNonRecursiveBinaryFilters}.
This result suggests that they can be efficiently used for other types of computations.
Their evaluation in the control of robotic systems with the Reservoir Computing framework is thus expected.

\section{Applications and results}
\zlabel{section:applications}
Most of the Reservoir Computing implementations previously discussed come from studies in the robotic field.
Those addressed problems and proposed solutions for different types of tasks.
A discussion on each of those applications and the role of Reservoir Computing in their resolution is in the scope of this review, and will help to understand the power and the weak points of this framework in real-world problems.

In this section, results obtained by Reservoir Computing implementations will be presented, discussed and, if possible, compared with the state-of-the-art approaches for the task.

\subsection{Pattern generations}
One of the most common task explored during experiments, consisted in the usage of the framework for learning specific periodic outputs.
This shows the ability of the system to deal with complex data and to adapt its behaviour to incoming cues.
The test is thus optimal for evaluation of proposed reservoirs.

Generating systems are also present in natural entities, where they take the name of Central Pattern Generators (CPGs) \cite{CentralPatternGenerator2000, CentralPatternGenerator2015}.
Those are fundamental building blocks for the control of gait patterns in humans and animals.
Indeed, their usage concur in the generation of smooth gait patterns.
These systems are networks composed of groups of neurons that generate signals used for step movements.
Feedback inputs are used for correcting the gait generation in case, for example, of uneven grounds \cite{CentralPatternGenerator2015}.
However, those systems do not rely on those cues.
They are capable of generating patterns also when sensory inputs are no longer available.
This makes the task hard in that it requires the system to correctly manage the cues while being able to autonomously generate the pattern.
Moreover, the lack of a rhythmic input requires the system to embed a time behaviour.

In natural entities, Central Pattern Generators produce signals that are used for the control of limbs and articulations.
Classical robotic systems follow the same principle, with the movement that is generated through the control of motors and joints.
These control signals have to be specialised in the control of each single actuator.
Moreover, they have to be synchronised between each other.
The combination of single, harmonic joint movements, contributes then in the generation of the robot gait.

Differently from other types of neural networks, reservoirs possess recurrent links that give the echo-state property.
This implicit awareness of time is used in the task for the generation of the periodic patterns.
Moreover, it theoretically allows the system to work also without input signals, mimicking natural behaviour \cite{OnTheDefinitionOfCentralPatternGeneratorAndItsSensoryControl} and embedding the timing properties.
This possibility is however difficult to obtain in that it requires endogenous cues\footnote{Signals generated from inside the system.} not to fade away with time. 
For this reason, a common approach consists in the exploitation of feedback links to keep the generated signal stable without the need of external inputs.

Despite their ability of input-less generation, Central Pattern Generator can obviously receive external cues for the tuning of the produced pattern.
To mimic this approach and to make the implemented solutions more useful, most of the implementations seen in the literature consider at least one type of input.
One of the approaches is to take sensory readings, using them as feedback of the action in the environment.
This mimics what commonly happens in natural systems.
Another approach is to provide control signals as inputs to the network.
This allows more complex operations to be executed.
Possible usages include the selection of different step types/behaviours \cite{DesignOfACentralPatternGeneratorUsingReservoirComputingForLearningHumanMotion, BehaviorSwitchingUsingReservoirComputingForASoftRoboticArm, ASoftBodyAsAReservoirCaseStudiesInADynamicModelOfOctopusInspiredSoftRoboticArm}.
Alternative usages comprehend frequency modulation of the generated signals and control of the gait speed \cite{PopulationsOfSpikingNeuronsForReservoirComputingClosedLoopControlOfACompliantQuadruped}.
This latter task shows to be harder to achieve in that it requires the reservoir to internalise a specific dynamic, being able to reach what in \cite{FrequencyModulationOfLargeOscillatoryNeuralNetworks} the authors call `equilibrium'.

In case of external controls, the generation has to face another problem.
The system has indeed to maintain a stable and robust behaviour also in case of noise or missing data.
Experiments about pattern generation with Reservoir Computing presented great capabilities in recovering from induced perturbations \cite{DesignOfACentralPatternGeneratorUsingReservoirComputingForLearningHumanMotion}.
Although often not able to correctly proceed during the occurrence of the error, the framework seems capable of successfully recover from it.
This seems to be true also when the error last for long time \cite{ASoftBodyAsAReservoirCaseStudiesInADynamicModelOfOctopusInspiredSoftRoboticArm}.
In this case however, a shift in the phase of the pattern is visible.
This fault recovery feature is important in that it allows a more robust behaviour of the generator.

An advantage on using a Reservoir Computing for pattern and gait generation lies in the possibility of effectively controlling soft bodies.
Those are composed of soft components that make the system highly dynamical and hard to control.
Differently from hard bodies, the state of the soft ones is influenced not only by internal actuators and properties, but also by the external environment.
This interaction with the space around can indeed change their state transition and thus their response to stimuli \cite{EnvironmentalandStructuralEffectsOnPhysicalReservoirComputingWithTensegrity}.
It is proposed that the pattern generation and emulation of non-linear dynamical systems, obtained from an octopus arm immersed in water, should not be related only to the dynamics of the body, but also to its interaction with the underwater environment \cite{ASoftBodyAsAReservoirCaseStudiesInADynamicModelOfOctopusInspiredSoftRoboticArm}.
The space around can thus enhance the non-linearity of the transformations, making the system harder to control but also possibly more useful for the Reservoir Computing approach.
One possibility consists in the exploitation of this relationship to automatically adapt the gait pattern to a specific type of terrain or environment.
It is thus possible to say that soft bodies are fully embodied in the space around, in that the state of one can perturb the state of the other \cite{AreRobotsEmbodied}.

Reservoir Computing capabilities perfectly cope with the soft bodies high dynamic.
The framework can indeed interact with them in two possibly overlapping modalities.
As first option, it can profit from their high non-linearity and intrinsic fading memory to use them as reservoirs; exploiting them to execute computations.
Alternatively, it can be used to learn the model of the body to improve its control capabilities; compared to classical approaches.
These two modalities are not disjoint, in that the same system can be used both for computation and as an actuator.
This approach is visible in \cite{PhysicalReservoirComputingWithOrigamiAndItsApplicationToRoboticCrawling}, where a soft body is used to produce a gait pattern for its own movement.
If effectively trained, an approach of this type can be used to cope with changes in the environment, adapting its behaviour to it.

Until now, the signal generation has been used for the control of hard and soft bodies' gait patterns.
However, mixed systems that combine flexible and stiff components exist.
Their control can thus exploit the computational power of the soft bodies, while simplify the overall system dynamics thanks to their hard parts.
This approach is followed in \cite{SpineDynamicsAsAComputationalResourceInASpineDrivenQuadrupedLocomotion}, where a robot with a stiff body and a flexible spine is proposed.
The backbone is then used to power the locomotion and, at the same time, embedded sensors are used to extract the system state used by the readout.
This mixed system showed good movement abilities, being able to carry loads until a third of its own body weight.

Simple control signals, like the ones that have been presented until now, can be used for the generation of constant gait patterns.
This approach implies that more complex outputs may produce more elaborate movements.
Implementations of the Reservoir Computing framework might then be able to learn advanced movements in addition to the classical ones.

\subsection{Drawing and motion}
Experiments trying to achieve the drawing goal seem strictly related to the Central Pattern Generator task.
Both of them require the system to generate an output that is used for a movement action.
Also this generation controls the single joints of the robot in a synchronised way.
The difference between the two tasks consists in the added complexity of the drawing application.
Central pattern generators have to deal mostly with timing of periodic patterns, while drawing conceptually focus on a spatial reasoning/computation.

Although many kinds of robots may be used for drawing shapes, the common approach is to use mechanical arms, that more closely mimic human actions.
Those types of robots add complexity compared to, for example, simple wheels-powered robots.
The latter move in a two-dimensional system; the other usually possess three degrees of freedom, requiring the medium to synchronise its movements in all the dimensions.
Investigations on the possible usage of Reservoir Computing for the control of this type of robot are interesting, in that those are the most common in the industrial environment.

The drawing task usually requires the robot to draw simple shapes like circles, triangles and squares.
This can also be associated to a movement task.
The difference between the two is that the usual solution for drawing requires the system to learn the trajectory to draw the shape; the motion task instead usually take the coordinates to reach with the movement. Basically, one can be modelled in a closed-loop fashion, while the other is usually defined in an open-loop one.

A novel approach to achieve the goal uses a system composed of parallel Liquid State Machines \cite{DiverseNoisyAndParallelANewSpikingNeuralNetworkApproachForHumanoidRobotControl}.
Each of them is trained to produce the same control outputs.
These are then averaged to obtain the final values.
The goal of this solution is to reduce errors of a single framework implementation.
Although good results were obtained, no rigorous comparison to a single-system is made, not showing if real improvements are brought to the final result.

Other approaches to the problem use Liquid State Machines with `dynamic synapses' \cite{ShortTermPlasticityInALiquidStateMachineBiomimeticRobotArmController}.
The aim is to give the system `short-term plasticity', a possibility already described in \cite{RealTimeComputingWithoutStableStatesANewFrameworkForNeuralComputationBasedOnPerturbations}.
However, not many subsequent implementations use this advanced version.
The lack of usage is probably related to the not agreeing results presented in \cite{AnExperimentalUnificationOfReservoirComputingMethods}.
Dynamic synapses do not seem to greatly increase the system performance and, additionally, they increase its complexity.
This observation is also confirmed in the drawing task, where their effectiveness is limited to just one shape over three.
Those results suggest that the utility of `dynamic synapses' may be limited to just few specific applications, and generally not to the drawing task.
Therefore, their usage should be carefully considered against the increasing complexity.

Finally, one approach used in \cite{MACOPModularArchitectureWithControlPrimitives}, consists in the control of the robot arm by mean or a hierarchical architecture, where multiple reservoir modules learns how to perform subsections of the task (called `control primitives' in the article).
This approach produces better results than same size single reservoir, probably because the task learned by each sub-module is easier.

\subsection{Direct kinematics and dynamics}
Direct kinematics allows to compute the position of the tip of the robotic arm (i.e., end-effector) from joints parameters.
Direct dynamics describes instead how the coordinates and velocities change due to forces and torques that act on the robot.
Those transformations are used to simulate how changes in control parameters influence the final state of the mechanical system.
Thanks to those predictions, it is possible to adapt the motion or the strategy to the context of the action.
These models are respectively the opposite of the inverse kinematics and dynamics, where the position and motion are used to learn the joints parameters and forces applied to the system.
An analytical approach is usually used for the computation of these transformations, but it requires the knowledge of the actuators characteristics.
Because of that, usage of learning techniques gained popularity.
Automatic learning model may be indeed applied to every kind of robotic body, which will automatically learn a model for the direct kinematic/dynamic transformation.
These techniques do not require any preliminary knowledge about the system.
Nevertheless, the state-of-the-art algorithms use sampled data as the basis for their calculation, such that their accuracy increases, so their complexity does.
The approach proposed in \cite{AReservoirComputingApproachForLearningForwardDynamicsOfIndustrialManipulators} aims to resolve this problem through the usage of a Reservoir Computing implementation: an Echo State Network with an input layer performing principal component analysis\footnote{Generalised Hebbian Learning algorithm (GHL), also called Sanger's rule}.
Although the reservoir should be able to decorrelate the inputs by itself, this choice may allow to reduce network size and complexity, since basic operations are performed in a previous, optimised layer.
The results show a better accuracy for the Reservoir Computing solution, together with a faster computing time.
Beside these conclusions, also a multi-step prediction is tested.
This uses the previously processed state of the robot to perform the subsequent prediction.
The accumulated error increases each step, and a single big failure might then affect the whole process.
This differs from the previous task in that the error is not reset after each single prediction.
The test requires then the system to be accurate, robust, and stable with respect to the expected result.
The authors introduced a leaky integrator with the aim to reduce the accumulated error.
The final results were indeed more accurate compared to the ones obtained by the `adversary' algorithms.

The previously described studies evaluates a system composed of an industrial robotic arm controlled by an Echo State Network.
The goal is for the system to internalise the dynamics, in order to predict the evolution in time of the state of the body.
A similar approach is proposed also in \cite{MorphologicalComputationBasedControlOfAModularPneumaticallyDrivenSoftRoboticArm}, for the prediction of the next position of a soft robotic arm used as a reservoir.
The system is composed of multiple Pneumatic Artificial Muscles (PAMs) and additional loads, which add non-linearity to the dynamics.
Also in this case of a high-dynamical system, the Reservoir Computing framework obtains good results.

\subsection{Terrain classification}
Terrain classification is an important task in that it allows the robot to adapt its gait to perform better on different soils.
It is also of complex solution, since it cannot be successfully detected through cameras.
The common way is thus to analyse sensors' readings, like the intensity of the ground contact.
Despite the availability of these data, most of the techniques for the task have problems classifying soft terrains.
Reservoir Computing seems instead to be able to provide good classification results also in this last type of soil.
This seems to be due to the framework ability to deal with sequences of data.
The analysis requires indeed the system to detect the response of the terrain when the contact occurs.
This would initially deform under the weight of the robot, eventually stabilizing.
The obtained information can be used to understand the physical characteristics of the soil.
Good results from the application of this technique are presented in \cite{HapticFeedbackwithAReservoirComputingBasedRecurrentNeuralNetworkForMultipleTerrainClassificatinOfAWalkingRobot}, where the system outperforms other classical approaches when soft soils, like sand and foam, are analysed.
Moreover, the system is able to compensate and predict missing data from sensors.

Compared to other methods, Reservoir Computing approaches seem also to suffer less accuracy degradation when fewer sensors are available \cite{TerrainClassificationForAQuadrupedRobot}.
This may denote that the temporal analysis carried on by the framework is able to extract information that would not normally be available for uncorrelated input data.
The extracted knowledge can thus be used to compensate the missing readings from the sensors.
This capability suggests that an approach exploiting the Reservoir Computing framework can lead to the production of more robust systems.
Alternatively, this information filling can be used to reduce the number of the sensors and, consequently, the power consumption and the complexity of the robot.

The sensors are not the only possibility to detect changes in the soil: the body itself may indeed allow an implicit analysis of the space around.
In \cite{TheBodyAsAReservoirLocomotionAndSensingWithLinearFeedback}, a tensegrity structure is used to generate a basic motion through oscillations.
Those kinds of system are known to be highly embodied, in that the change of environmental characteristics can easily modify their behaviour.
This implicit sensitivity is used by the authors to train a specific readout network.
The obtained classifier is then able to detect changes in the oscillation, and thus identify different types of terrains.
The final outcome basically uses the tensegrity structure itself as a sensor.
It has to be noted that this experiment is executed in a simulated environment, making this approach probably not reliable in real life systems.
Nevertheless, this technique uses the changes in the system itself, showing that it is possible to perceive the environment without the need of sensors.
Finally, this approach shows an uncommon possibility that may lead to future investigations.

\subsection{Localization and navigation}
Localization is a complex problem in robotics that consists in the inference of the position of the robot in the space.
This usually requires specific sensors and a knowledge of the environment to be exploited.
The model of the space may be provided from the start or may be discovered by the robot itself through exploration.
The storage of the space's characteristics may represent a problem both for the format of the memorization and for the size of the model itself.
This may be represented through multiple techniques, spacing from graphs and grids of navigable areas, to a direct representation of the objects of the environment.

Although a model is needed for the localization task, it also requires adequate sensors to be used.
Positioning systems (GPS, GLONASS, Galileo) and landmarks are the classical approaches, while odometry does not necessarily need sensors to be used.

While a coordinate system is enough to represent position, a knowledge of the space may be useful.
Recognizing features of the environment may enhance location inference, and it may be used when the positioning system is not available or it is not working properly.
Moreover, a model of the environment is useful for the navigation task.

The navigation consists in moving the robot from its location to a given destination.
This task obviously benefits from a clear space representation in that it may generate an optimised path, avoiding dead-ends routes.
The solution to the problem is fundamental for many real life tasks, like moving an object in a robotised environment.
It should be clear why the studies on the topic are so various in literature.

Common solutions and approaches require explicit models of the environment or specific sensors.
However, recent experiments tried to achieve those tasks through Artificial Neural Networks, avoiding to explicitly define a model or to equip the robot with specific hardware.
Reservoir Computing systems shown properties useful for those tasks.

A basic approach, often used in literature for indoor localization, consists in the usage of Radio Signal Strength (RSS).
This solution evaluates the signal intensity from fixed beacons, with the aim of triangulating the position.
The approach uses cheap sensors but, on the other hand, the results are not very precise, with error rates of almost one meter.
In \cite{RSSbasedRobotLocalizationInCriticalEnvironmentsUsingReservoirComputing} the authors train an Echo State Network with leaky integrator, with the aim of obtaining more precise results.
The final error is way smaller than the typical one.

Another proposed solution tries to mimic the brain localization system of rats.
This consists of an area of the hippocampus whose neurons activate when the animal is in a specific location.
Basically, the knowledge of the previously visited areas and the perception of environmental characteristics (allothetic inputs) bring to the activation of specific neurons that represent the position in the space \cite{TheHippocampusAsASpatialMapPreliminaryEvidenceFromUnitActivityInTheFreelyMovingRat}.
This is also influenced by the pure movement of the rat (idiothetic inputs): an approach used also by the odometry solutions.
Recurrent Neural Networks are able to mimic this behaviour.
This is possible through their fading memory property that provides the recent history of the movement.
Unifying the knowledge of the past states with the features perceived from the environment, it is possible to obtain spatio-temporal patterns to be used to create a robust positioning system.
This setup is completely represented by the Reservoir Computing framework, where the readout is trained for obtaining the position from the patterns.
This is done in \cite{EventDetectionAndLocalizationForSmallMobileRobotsUsingReservoirComputing}, where the readout is trained as a classifier to identify the position on a grid of previously defined areas.
Those are fixed and a specific neuron activation is associated to each of them.
In the discussed paper the environment is known.
Because of that, an offline training method is used.
Nevertheless, the framework may be used also for unsupervised learning, as shown in \cite{LearningSlowFeaturesWithReservoirComputingForBiologicallyInspiredRobotLocalization}.
In this case, only the number of different areas is known, and the robot has to learn how to distinguish between them.

Those experiments show good results in the usage of Reservoir Computing framework, resulting in the robot being able to recognise different areas without the need of specific sensors.
Indeed, the only input received is composed of distance readings, showing the ability of the network to infer actual state as a combination of environmental data and past history.
The need of a sort of memory is confirmed by the fact that the robot is able to infer its position in identical but distinct areas.
This behaviour is possible only if the system is able to track back to the previous states (i.e., visited locations).
Also if some errors are present, especially in the unsupervised experiment, the overall results are satisfying.
Moreover, as already shown in other tasks, the reservoir seems to be able to correct erroneous belief and predictions.
This is shown in experiments considering dynamic obstacles moving in the environment.
The errors obtained in the localization task are obviously greater, but the framework manages to obtain acceptable results and/or to correct erroneous predictions.
This also confirms the robustness of Reservoir Computing techniques.

For the localization task, a choice carried on in \cite{LearningSlowFeaturesWithReservoirComputingForBiologicallyInspiredRobotLocalization} is to use a `queue' of layers.
Specifically, the first performs a `slow features analysis' with the aim to filter fast changing information.
Those may be related to noise or dynamic elements moving in the environment, and thus perturbing the search of reference points.
The second network performs instead the effective area classification with the data `filtered' by the previous layer.

The studies about localization tasks suggest that Reservoir Computing is able to extract important information from the input and build an implicit map of the environment.
This is also confirmed in \cite{GenerativeModelingOfAutonomousRobotsAndTheirEnvironmentsUsingReservoirComputing} where the `knowledge' acquired by the system is used to build an explicit map.
Nevertheless, some investigations are still needed, like finding a relation between the size of the network and the quantity of data that it can store.
Moreover, studies about the possibility of using online methodologies, dynamic environments and test the persistence of the implicit map during time are still in a beginning state.

To conclude, in \cite{EventDetectionAndLocalizationForSmallMobileRobotsUsingReservoirComputing}, while performing navigation, also an event detection task is tested.
Here the robot has to identify events like passing through a door.
The events are inferred by a readout trained as a classifier.
Although not directly related with the localization task, this experiment exploits its same principle and methodology, still providing good results.

\subsection{Collision avoidance and target capture}
Collision avoidance is usually considered a fundamental capability in the robotic field.
Although being often easily achieved, its experimentation may be useful in the evaluation of new types of systems.
This is true in the task of evaluating biological networks as reservoirs, as their usage is still an emergent area of research.
Indeed, although their computational abilities are unquestioned, their effectiveness in usage with the actual technologies is still to be assessed.
Basic tasks may be used for assessing their potential with the technologies actually available.

The few studies on the usage of Reservoir Computing with biological networks, start with the task of generating a controlled output signal.
In \cite{ReservoirComputingWithDissociatedNeuronalCulture}, the system initially learns how to generate a coherent temporal pattern.
It is tested as controller of a robot in the task of reaching a goal while avoiding collisions.
The network input are perturbations from sensors, and the direction is decided from the error between the output signal and the target function.
The authors claim that the system is able to achieve the task.
Different results are instead obtained in \cite{TowardsMakingACyborgACloseLoopReservoirNeuroSystem}, where a biological system is used for the control of a robot motors.
As for the previous study, the first step consists in the generation of a temporal pattern, with the additional ability to change the signal frequency for a brief period of time after a stimulation.
This capability suggests the presence of a fading-memory property.
The experiment continues with the collision avoidance task, showing however mediocre results, with the robot simply going in circle.

Beside its usage for the evaluation of Biological Neural Networks, collision avoidance has been used in combination with the `seek and capture' task to evaluate the capability of a Reservoir Computing implementation to learn two conceptually opposite tasks.
While the first requires indeed to dodge obstacles, the second consists instead in reaching them.
In \cite{ModelingMultipleAutonomousRobotBehaviorsAndBehaviorSwitchingWithASingleReservoirComputingNetwork}, the authors show that the learning of two opposite types of tasks is possible.
Indeed, the reservoir does not directly learn the tasks, but it only contains and elaborate the information obtainable from the sensors and from the control inputs.
The exploitation of these data is made by the readout, that is then trained to associate the output of the reservoir in order to obtain the correct behaviour.
This capability and concept allow to learn opposite tasks in an easy way.
Behaviour switching is however done through an external command.
An improved version of the task trains each behaviour in a different network, finally using an upper layer reservoir to merge the previously generated control signals \cite{ModularReservoirComputingNetworksForImitationLearningOfMultipleRobotBehaviors}.
The resulting trajectory smoothly switches between the two behaviours, vying with a vector-based mixing method.
The major downside of the architecture is its reliance on a big mixing reservoir, that raises the doubt that the final system might be oversised for the required task.
A similar approach is used in \cite{SupervisedLearningOfInternalModelsForAutonomousGoalOrientedRobotNavigationUsingReservoirComputing}, where two different Reservoir Computing systems are used in a layered fashion to achieve the previously described goals.

A task of navigation with the aim of finding and capture a coloured ball is presented in \cite{UsingEchoStateNetworksForRobotNavigationBehaviorAcquisition}.
The approach consists in the usage of a camera connected to the reservoir as input.
The used learning data are captured directly by a human execution of the task, and thus containing biases such as the knowledge of the position of the ball also when it is not visible to the robot.
The system is however able to achieve the goal with success, showing ability in the analysis of the spatial data from the camera.
Nevertheless, it has to be noted that the task is not particularly hard in that it does not contain any time related requirement.
Indeed, a multi-layer perceptron also obtained good results, although worse compared to the ones obtained by the reservoir system.
The process of collecting data from a human controlled system is followed also in \cite{OctopusInspiredSensorimotorControlOfAMultiArmSoftRobot}.
The robot uses a six-arms motion apparatus to reach a balloon.
As for the previous task, the system uses the images obtained from a camera for the seek part.
The capture part only approaches the target, decreasing the speed as the robot got near to it.
The added complexity of the task is just the speed control: the limbs used for the movement are indeed controlled externally, through the directions generated by the readout.

Finally, in the context of robot gait control, Reservoir Computing has been also used as a sort of sensor to detect environmental features and chose the correct gait pattern automatically.
This approach is used to pass over holes in the ground, basically applying something similar to a three-dimensional avoidance \cite{ReservoirBasedOnlineAdaptiveForwardModelsWithNeuralControlForComplexLocomotionInAHexapodRobot} \cite{DistributedRecurrentNeuralForwardModelsWithSynapticAdaptationAndCPGBasedControlForComplexBehaviorsOfWalkingRobots}.
The approach uses the predicted feedback of each leg to detect gaps in the ground and react accordingly, changing the gait pattern.
The last step is however achieved by an external controller, in that the reservoir is only used to detect the obstacle/hole.

\section{Conclusion}
\zlabel{section:conclusion}
This review summarises the recent most interesting approaches of Reservoir Computing in the robotic field.
Initially, the framework idea and its variant implementations are presented.
This results in the identification of four main derived architectures.
Subsequently, the reservoirs are classified according to their overall characteristics and properties.
The resulting classes show to be intersected, with some systems belonging to more than one.

The literature about physical reservoir (section `\ztitleref{section:physical}') is wide, but regarding the robotic field no interesting result is found.
Despite lack of articles, this is one of the most interesting and promising branch of study, with physical reservoirs being known for low-power consumption and high-speed computation.

Similarly, just few solutions analysing biological systems (section `\ztitleref{section:biological}') are found.
These results also seem far to be successfully exploited in robotics, mostly due to their intrinsic limitations in production and the lack of appropriate technologies.

Literature about morphological computing bodies (section `\ztitleref{section:morphological}') is instead wide and active, with many studies trying to exploit those systems with the Reservoir Computing approach.
The results are however not always satisfying, suggesting that their usage greatly depends on the characteristics of the reservoir.
Moreover, a preliminary evaluation of the system computational capabilities is often not possible.

Finally, literature about software reservoir (section `\ztitleref{section:software}') is wide, and presents many interesting results obtained in multiple tasks.
With this type of system, Reservoir Computing framework sometimes achieves better results than the `state-of-the-art' approaches.

An overview of applications and experiments concerning Reservoir Computing is carried on (section `\ztitleref{section:applications}').
In all the macro-categories representing tasks, the framework performs well and often better that classical/state-of-the-art approaches (e.g.
control of soft bodies).
In others, the performance are worse but more robust against noise and/or missing data.
Moreover, solutions using the Reservoir Computing framework are often able to work with cheaper sensors and hardware.

Reservoir Computing shows to be a valid approach, and many studies in the field present good results in its usage.
Experiments in the robotic area are however mostly limited to the software systems, lacking advanced evaluation of more `exotic' reservoirs.
Future improvements may approach this latter area of study, mostly thanks to development of physical and morphological reservoirs.
Those are indeed the most promising type of systems, possibly allowing improvements in power consumption, computational power and cost-effectiveness.
Increasing attention towards biological systems is also possible, but due to the cost of the tools and the multi-disciplinary knowledge needed, its approach may be slower.
Undoubtedly, development of Reservoir Computing will facilitate future improvements of robotic research, possibly allowing to get closer to the creation of real `intelligent' robots.

\printbibliography

\end{document}